\documentclass[11pt]{article}
\usepackage{coling2020}
\usepackage{times}
\usepackage{url}
\usepackage{latexsym}

\usepackage{graphicx}
\usepackage{float}
\usepackage{amsmath}
\usepackage{amssymb}
\usepackage{multirow}
\usepackage{booktabs}
\usepackage{url}
\usepackage{array}
\usepackage{enumitem}
\usepackage{algorithm}
\usepackage{algorithmic}
\usepackage{pifont}
\usepackage{CJKutf8}

\title{A Sentence Cloze Dataset for Chinese Machine Reading Comprehension}

\author{Yiming Cui$^{\dag\ddag}$, Ting Liu$^\dag$, \bf Ziqing Yang$^\ddag$, Zhipeng Chen$^\ddag$, Wentao Ma$^\ddag$, \\ 
{\bf Wanxiang Che$^\dag$, Shijin Wang$^\ddag$$^\S$, Guoping Hu$^\ddag$} \\
{$^\dag$Research Center for SCIR, Harbin Institute of Technology, Harbin, China} \\
{$^\ddag$State Key Laboratory of Cognitive Intelligence, iFLYTEK Research, China} \\
{$^\S$iFLYTEK AI Research (Hebei), Langfang, China} \\
{$^\dag$\tt \{ymcui,tliu,car\}@ir.hit.edu.cn} \\
{$^\ddag$$^\S$\tt\{ymcui,zqyang5,zpchen,wtma,sjwang3,gphu\}@iflytek.com} \\  
}

\date{}

\begin{document}
\begin{CJK*}{UTF8}{gbsn}
\maketitle

\begin{abstract}
Owing to the continuous efforts by the Chinese NLP community, more and more Chinese machine reading comprehension datasets become available.
To add diversity in this area, in this paper, we propose a new task called Sentence Cloze-style Machine Reading Comprehension (SC-MRC).
The proposed task aims to fill the right candidate sentence into the passage that has several blanks.
We built a Chinese dataset called CMRC 2019 to evaluate the difficulty of the SC-MRC task.
Moreover, to add more difficulties, we also made fake candidates that are similar to the correct ones, which requires the machine to judge their correctness in the context.
The proposed dataset contains over 100K blanks (questions) within over 10K passages, which was originated from Chinese narrative stories.
To evaluate the dataset, we implement several baseline systems based on the pre-trained models, and the results show that the state-of-the-art model still underperforms human performance by a large margin.
We release the dataset and baseline system to further facilitate our community.
Resources available through \url{https://github.com/ymcui/cmrc2019}
\end{abstract}

\section{Introduction}
Machine Reading Comprehension (MRC) is a task to comprehend given articles and answer the questions based on them, which is an important ability for artificial intelligence.
The recent MRC research was originated from the cloze-style reading comprehension \cite{hermann-etal-2015,hill-etal-2015,cui-etal-2016}, which requires to fill in the blank with a word or named entity, and following works on these datasets have laid the foundations of this research \cite{kadlec-etal-2016,cui-acl2017-aoa,dhingra-etal-2017}. 
Later on, SQuAD \cite{rajpurkar-etal-2016} was proposed, and the answer transformed from a single word to a span, which has become a representative span-extraction dataset and massive neural network approaches \cite{wang-and-jiang-2016,xiong-etal-2016,rnet-2017,mnemonic-2018,slqa-2018,qanet-2018} have been proposed which further accelerated the MRC research. 

\begin{figure}[tp]
\centering\scriptsize
        \begin{tabular}{p{0.48\textwidth} p{0.48\textwidth}}
        \toprule
	{\bf [Passage]} & {\bf [Passage]} \\ 
	"森林里有一棵大树，树上有一个鸟窝。{\bf [BLANK1]}，还从来没有看到过鸟宝宝长什么样。 
            小松鼠说：“我爬到树上去看过，鸟宝宝光溜溜的，身上一根羽毛也没有。” “我不相信，”小白兔说，“所有的鸟都是有羽毛的。” 
            “鸟宝宝没有羽毛。”小松鼠说，“你不信自己去看。” 
            小白兔不会爬树，它没有办法去看。小白兔说：“我请蓝狐狸去看一看，我相信蓝狐狸的话。” 小松鼠说：“蓝狐狸跟你一样，也不会爬树。” 
            蓝狐狸说：“我有魔法树叶，我能变成一只狐狸鸟。” {\bf [BLANK2]}，一下子飞到了树顶上。 “蓝狐狸，你看到了吗？”小白兔在树下大声喊。 
            “我看到了，鸟窝里有四只小鸟，他们真是光溜溜的，一根羽毛也没有。”蓝狐狸说。 就在这时候，鸟妈妈和鸟爸爸回来了，
            {\bf [BLANK3]}，立刻大喊大叫： “抓强盗啊！抓强盗啊！强盗闯进了我们家里，想偷我们的孩子！” 
            {\bf [BLANK4]}，全都飞了过来。他们扇着翅膀，朝蓝狐狸冲过来，用尖尖的嘴啄他，用爪子抓他。 蓝狐狸扑扇翅膀，赶紧飞。 
            鸟儿们排着队伍，紧紧追上来。{\bf [BLANK5]}，它飞得不高，也飞得不快。 “救命啊，救命！”蓝狐狸说，“我不是强盗，我是蓝狐狸！”  &
       A long time ago, there was a queen. {\bf [BLANK1]} Soon after the child was born, the Queen died. {\bf [BLANK2]} The stepmother didn't like her very much. She made Snow White do the housework all day and all night. A wizard had given this Queen a glass. The glass could speak. It was on the wall in the Queen's room. Every day the Queen looked in the glass to see how beautiful she was. As she looked in the glass, she asked: "Tell me, glass upon the wall, who is most beautiful of all?" And the glass said: "The Queen is most beautiful of all.". Years went by. Snow-white grew up and became a little girl. Every day the Queen looked in the glass and said, "Tell me, glass upon the wall, {\bf [BLANK3]}" And the glass said, "Snow-white is most beautiful of all.". When the Queen heard this, {\bf [BLANK4]}. She said, "Snow-white is not more beautiful than I am. There is no one who is more beautiful than I am.". So she called a hunter and said, "Take Snow-white into the forest and kill her.". The hunter took Snow-white to the forest, but he did not kill her, because she was so beautiful and so lovely. He put Snow White in the forest and went away.  \\
        \midrule
        {\bf [Candidates]} & {\bf [Candidates]}  \\ 
        0: 蓝狐狸是第一次变成狐狸鸟 & 0: The king married another queen \\
        1: 森林里所有的鸟听到喊声 & 1: She had a pretty daughter named Snow White \\
        2: 他们看到鸟窝里蹲着一只蓝色的大鸟 & 2: \underline{The king was also passed away} \\
        3: 蓝狐狸真的变成了一只蓝色的大鸟 & 3: who is most beautiful of all? \\
        4: 小动物们只看到过鸟妈妈和鸟爸爸在鸟窝里飞进飞出 & 4: \underline{she was very happy} \\
        5: \underline{小松鼠变成了一只蓝色的大鸟} & 5: she was very angry \\
        \midrule
        {\bf [Answers]} & {\bf [Answers]} \\ 
        4, 3, 2, 1, 0 & 1, 0, 3, 5 \\
        \bottomrule
        \end{tabular}
\caption{\label{cmrc2019-example} Examples of the proposed CMRC 2019 dataset. The candidate with underline means it is a fake candidate (does not belong to any blank). For clarity, we also provide an English example.} 

\end{figure}

Besides the MRC in English text, we have also seen rapid progress on Chinese MRC research.
\newcite{cui-etal-2016} proposed the first Chinese cloze-style reading comprehension dataset: People Daily \& Children's Fairy Tale (PD\&CFT). 
Later, \newcite{cmrc2017-dataset} proposed another dataset for CMRC 2017, which is gathered from children's reading books, consisting of both cloze and natural questions.
\newcite{he-etal-2018-dureader} proposed a large-scale open-domain Chinese reading comprehension dataset (DuReader), which consists of 200k queries annotated from the user query logs on the search engine.
In span-extraction MRC, \newcite{cui-emnlp2019-cmrc2018} proposed CMRC 2018 dataset for Simplified Chinese, and \newcite{shao2018drcd} proposed DRCD dataset for Traditional Chinese, similar to the popular dataset SQuAD \cite{rajpurkar-etal-2016}.
\newcite{zheng-etal-2019-chid} proposed a large-scale Chinese idiom cloze dataset.

Though various efforts have been made, most of these datasets stop at token-level or span-level inference, which neglect the importance of long-range reasoning of the context. 
Moreover, powerful pre-trained models such as BERT \cite{devlin-etal-2019-bert}, XLNet \cite{yang2019xlnet}, RoBERTa \cite{liu2019roberta} have surpassed human performance on various MRC datasets, such as SQuAD \cite{rajpurkar-etal-2016}, SQuAD 2.0 \cite{rajpurkar-etal-2018-know}, CoQA \cite{reddy2019coqa}, RACE \cite{lai-etal-2017}, etc.

To further test the machine comprehension ability, In this paper, we propose a new task called Sentence Cloze-style Machine Reading Comprehension (SC-MRC).
The proposed task preserves the simplicity of cloze-style reading comprehension but requires sentence-level inference when filling the blanks.
Figure \ref{cmrc2019-example} shows an example of the proposed dataset. 
We conclude our contributions in three aspects.
\begin{itemize}
	\item We propose a new machine reading comprehension task called Sentence Cloze-style Machine Reading Comprehension (SC-MRC), which aims to test the ability of sentence-level inference.
	\item We release a challenging Chinese dataset CMRC 2019, which consists of 100K blanks, to evaluate the SC-MRC task.\footnote{The data was used in the shared task of CMRC 2019 workshop, as thus, we directly name this dataset as CMRC 2019.}
	\item Experiments on several state-of-the-art Chinese pre-trained language models show that there is still much room for these models to surpass human performance, indicating that the proposed data is challenging.
\end{itemize}

\section{The Proposed Dataset}\label{dataset}
\subsection{Task Definition}
Generally, the reading comprehension task can be described as a triple $\langle \mathcal P, \mathcal Q, \mathcal A \rangle$, where $\mathcal P$ represents {\bf P}assage, $\mathcal Q$ represents {\bf Q}uestion, and the $\mathcal A$ represents {\bf A}nswer. 
Specifically, for sentence cloze-style reading comprehension task, we select several sentences in the passages and replace with special marks (for example, {\tt [BLANK]}), forming an incomplete passage. 
The sentences are identified using LTP \cite{che2010ltp}, and we further split the sentence with comma and period mark, as some of the sentences are too long.
The selected sentences form a candidate list, and the machine should fill in the blanks with these candidate sentences to form a complete passage.
Note that, to add more difficulties, we could also add the fake candidates, which do not belong to any blanks in the passage.

\subsection{Passage Selection}
The raw material of the proposed dataset is from children's books, containing fairy tales and narratives, which is the proper genre for testing the sentence-level inference ability, requiring the correct sentence order of the stories.
During the passage selection, we restrict the character-level passage length in the range of 500 to 750.
If the passage is too short, then there will be only few blanks in the passage. 
If the passage is too long, it will be harder for the model to process.
After the passage selection, we got 10k passages and split them into three parts for generating the training, development, and test set.

\begin{table*}[htpb]
\small
\begin{center}
\begin{tabular}{l c c c c c}
\toprule
 & \bf Genre & \bf Query Type & \bf Answer Type & \bf Doc \# & \bf Query \# \\
\midrule
PD\&CFT \cite{cui-etal-2016} 	& News, Story	& Cloze & Word & 28K & 100K \\
WebQA \cite{Li2016DatasetAN}	& Web & NQ & Entity & - & 42K \\
CMRC 2017 \cite{cmrc2017-dataset}					& News & Cloze\&NQ & Word & - & 364K \\
DuReader \cite{he-etal-2018-dureader}		& Web & NQ & Free Form & 1M  & 200K \\
CMRC 2018 \cite{cui-emnlp2019-cmrc2018}	& Wiki & NQ & Passage Span & - & 18K \\
DRCD \cite{shao2018drcd}		& Wiki & NQ & Passage Span & - & 34K \\
C$^3$ \cite{Sun2019ProbingPK}	& Mixed & NQ & Choices & 14K & 24K \\
ChID \cite{zheng-etal-2019-chid}  & News, Novels, etc. & Cloze & Chinese Idioms & 580K & 729K \\
\midrule
CMRC 2019					& Story & Cloze & Sentence & 10K & 100K \\
\bottomrule
\end{tabular}
\caption{\label{chinese-mrc} Comparisons of Chinese MRC datasets. NQ represents natural questions.}
\end{center}
\end{table*}

\subsection{Cloze Generation}
Sentence cloze task does not require human annotation as it only requires the selection of the blanks, and they will naturally become the answers.
However, to ensure a high-quality cloze generation, the following rules are applied.
\begin{itemize}
\setlength{\itemsep}{5pt}
\setlength{\parsep}{0pt}
\setlength{\parskip}{0pt}
	\item The first sentence is always skipped, which usually contains important topic information.
	\item Select the sentence based on the comma or period mark, resulting in the range of 10 to 30 characters. Note that we eliminate the comma or period at the end of the candidate sentence.
	\item If a part of a long sentence is selected, we do not choose other parts to avoid too many consecutive blanks.
\end{itemize}

\subsection{Fake Candidates}
In order to bring difficulties in this task and better test the ability of machine reading comprehension, we propose to add fake candidates to confuse the system. 
In this way, the machine should not only generate the correct order of the candidate sentences but also should identify the fake candidates that do not belong to any passage blanks.
A good fake candidate should have the following characteristics.
\begin{itemize}
\setlength{\itemsep}{5pt}
\setlength{\parsep}{0pt}
\setlength{\parskip}{0pt}
	\item The topic of the fake candidate should be the same as the passage.
	\item If there are named entities in the fake candidates, it should also appear in the passage.
	\item It could NOT be a machine-generated sentence, or it would be very easy for the machine to pick the fake one out.
\end{itemize}

A natural way to generate fake candidates is to adopt human annotation, while it is rather time-consuming.
In order to minimize the cost by human annotation, in this paper, we propose a novel approach to generate fake candidates that is qualified for the requirements above.

Typically, a complete story is rather long that we must truncate for easy processing by the machines.
In this context, we could directly pick the sentences outside the truncated passage within the same story.
As these sentences are still from the same story, the topic and name entities are in accordance with the main passage. 
Also, it is a part of the original story, which is a natural sentence rather than a machine-generated sentence.
Using the strategies above, we could generate many fake candidates and mix them with the correct candidates to form the final candidate sentences.

\subsection{Statistics}
The general statistics of the final data are given in Table \ref{data-stats}, and comparisons with other Chinese MRC datasets are shown in Table \ref{chinese-mrc}.
As we can see, the proposed dataset mitigates the absence of sentence-level inferential reading comprehension dataset. 
Note that the training set does not contain any fake candidates, as we want to test the generalization of the machine reading comprehension system without training on both real and fake candidates.

\begin{table}
\small
\centering
\begin{tabular}{l rrr}
\toprule
 & \bf Train & \bf Dev & \bf Test \\
\midrule
Context \# & 9,638 & 300 & 500 \\
Blank \# & 100,009 & 3,053 & 5,118  \\
Max Context Tokens \# & 731 & 717 & 717 \\
Avg Context Tokens \# & 642 & 632 & 633 \\
Max Candidate \# & 15 & 15 & 15 \\
Avg Candidate \# & 10.4 & 13.3 & 13.4 \\
Max True Candidate \# & 15 & 14 & 14 \\
Avg True Candidate \# & 10.4 & 10.2 & 10.2 \\
Max Candidate Tokens \# & 29 & 29 & 29 \\
Avg Candidate Tokens \# & 13.7 & 14.1 & 14.2 \\
\bottomrule
\end{tabular}
\caption{\label{data-stats} Statistics of CMRC 2019.}
\end{table}

\section{Baseline System}
In this paper, we mainly adopt BERT and its related variants for our baseline systems.
\begin{itemize}
	\item {\bf Input Sequence: } Given a passage $p$ and its $n$ answer options $\{a_1, a_2, \ldots, a_n\}$, we first replace the blanks in $p$ with the special tokens {\tt[unused{\it Num}]} from the vocabulary to fit the input format of BERT, where {\it Num} ranges from $0$ to number of blanks $-1$. Then for each $a_i$ in the answer options, we concatenate $a_i$  and $p$ with the token {\tt[SEP]} as the input sequence.
	\item {\bf Main Model: } The input sequence of length $l$ is fed into BERT to get the hidden representations $H\in \mathbb{R}^{l\times d}$. The dot product of $H$ with trainable parameters $w\in \mathbb{R}^d$ gives the logits $t=H\cdot w$, where $ t\in \mathbb{R}^{l}$. Finally, the probabilities of the blanks for the current option is calculated by a softmax over the logits with only positions of blanks unmasked. The training objective is to minimize the cross-entropy between the predicted probabilities and the ground-truth positions. 
	\item {\bf Decoding: } The model outputs the predictions for answer options in terms of the probabilities of blanks they can be filled into. They need to be transformed into the predictions for blanks in terms of the answers they choose. A simple method we used is, among all the answer options for a passage, taking the option that gives the highest probability to a blank as the prediction for that blank (each option is allowed to be the prediction of multiple blanks).
\end{itemize}

\section{Experiments}
\subsection{Evaluation Metrics}
We adopt two metrics to evaluate the systems on our datasets, namely Question-level Accuracy (QAC) and Passage-level Accuracy (PAC).
QAC is calculated by the ratio between the correct predictions and total blanks.
Similarly, PAC is to measure how many passages have been correctly answered. 
We only count the passages that all blanks have been correctly predicted.
\begin{gather}
\mathbf{QAC} = \frac{\text{\# correct predictions}}{\text{\# total blanks in dataset}} \times 100\% ~~;~~
\mathbf{PAC} = \frac{\text{\# correct passages}}{\text{\# total passages in dataset}} \times 100\%
\end{gather}

\subsection{Experimental Setups}
We adopt Chinese BERT-base, BERT-multilingual, Chinese BERT-wwm and RoBERTa with whole word masking \cite{chinese-bert-wwm,cui-2020-revisiting} as backbones.
Note that both genres share the same vocabulary of WordPiece \cite{wu2016google} tokens as the same in Chinese BERT\footnote{https://github.com/google-research/bert/blob/master/multilingual.md}, which have 21,128 words. 
All models are trained with 3 epochs on Tesla V100, with an initial learning rate of 3e-5, a maximum sequence length of 512, and a batch size of 24.
The implementation was done on PyTorch \cite{paszke2017automatic} with Transformers library \cite{Wolf2019HuggingFacesTS}.

\subsection{Results}
The baseline results are shown in Table \ref{result-baseline}. 
As we can see, the Chinese BERT-base model could give a QAC of 71.2 and 71.0 on the development and test set, respectively.
However, with respect to the PAC metric, it only gives an accuracy of below 10, which suggests that there is plenty of room for optimizing the sentence cloze procedure to consider not only the single cloze but also the coherence of the whole passage.
BERT with whole word masking strategy substantially outperform original BERT implementation, and using the large model could also give a significant boost on both QAC and PAC metrics.

To evaluate human performance, we invited qualified annotators (English-majored students) to solve the sentence clozes of 100 passages in the development and test set (randomly sampled), respectively, resulting in 1,016 and 1,027 blanks for each. 
For each set, three annotators are involved.
Then we calculate the average QAC and PAC to roughly estimate the human performance on this dataset.

Comparing the RoBERTa-wwm-ext-large with the human performance, though there is only a gap of 13.3 on QAC, there is a significant gap on PAC, which also suggests that more attention should be drawn on the accuracy of the passage as a whole. 
We also include the top systems in our evaluation campaign, which used various approaches for improving the final performance, including pseudo-training data generation, data augmentation, ensemble, etc.
However, comparing these models with human performance, we can see that there is still much room for improvement, indicating that our dataset is challenging.

\begin{table}[htpb]
\small
\centering
\begin{tabular}{p{4cm}  c c c c}
\toprule
\multirow{2}*{\bf System} & \multicolumn{2}{c}{\bf Dev} & \multicolumn{2}{c}{\bf Test} \\
& \bf QAC & \bf PAC & \bf QAC & \bf PAC \\
\midrule
{\em Human Performance}	& \em 95.9 & \em 81.0 & \em 95.3 & \em 75.0 \\
Random Selection			& 7.6 & 0.0 & 7.5 & 0.0 \\
\midrule
\multicolumn{5}{l}{\em Top Submissions from CMRC 2019} \\
bert\_scp\_spm$^\dag$  	& 90.9 & 60.0 & 90.8 & 57.6 \\
mojito$^\dag$  			& 88.2 & 48.0 & 86.0 & 41.8 \\
DA-BERT$^\dag$			& 86.3 & 34.3 & 84.4 & 27.6 \\
\midrule
\multicolumn{5}{l}{\em Baseline Systems} \\
BERT			 		& 71.2 & 10.0 & 71.0 & 8.8 \\
BERT-multilingual 		& 66.8 & 6.67  & 66.0 & 5.0 \\
BERT-wwm				& 72.4 & 9.3 & 71.4 & 7.6 \\
BERT-wwm-ext				& 75.0 & 12.7 & 73.7 & 9.2 \\
RoBERTa-wwm-ext		 	& 75.9 & 11.0 & 75.8 & 12.4 \\
RoBERTa-wwm-ext-large	& 82.6 & 23.3 & 81.7 & 23.0 \\
\bottomrule
\end{tabular}
\caption{\label{result-baseline} Experimental results on CMRC 2019. The ensemble system (unpublished) marked with $\dag$. }
\end{table}

\section{Conclusion}\label{conclusion}
In this paper, we proposed a new task called Sentence Cloze-style Machine Reading Comprehension (SC-MRC) and released a Chinese dataset for evaluating the sentence-level inference ability.
The proposed dataset contains both real and fake candidate sentences for filling the clozes, which not only requires the machine to choose the correct sentence but also distinguishes the real sentence from fake sentences.
We built up baseline models based on the popular pre-trained language models, and the results show that the state-of-the-art models still underperform the human performance, especially on PAC evaluation metric.

We hope the release of this dataset could bring language diversity in machine reading comprehension tasks and accelerate further investigation on solving the questions that need comprehensive reasoning over multiple clues.

\section*{Acknowledgments}\label{ack}
We would like to thank all anonymous reviewers for their thorough reviewing and providing constructive comments to improve our paper. 
This work was supported by the National Natural Science Foundation of China (NSFC) via grant 61976072, 61632011, and 61772153.

\bibliography{coling2020}
\bibliographystyle{acl}

\end{CJK*}
\end{document}